\documentclass[letterpaper]{article} 
\def\showauthors@on{T}
\usepackage[submission]{aaai2026}  
\usepackage{times}  
\usepackage{helvet}  
\usepackage{courier}  
\usepackage[hyphens]{url}  
\usepackage{graphicx} 
\usepackage{natbib}  
\usepackage{caption} 
\usepackage{algorithm}
\usepackage{algorithmic}

\usepackage{newfloat}
\usepackage{listings}

\usepackage{booktabs}
\usepackage{multirow}
\usepackage{amsmath}
\usepackage{cleveref}
\usepackage{xcolor}
\usepackage{colortbl}

\definecolor{tableheadcolor}{rgb}{1, 1, 1}
\definecolor{lightgray}{rgb}{1, 1, 1} 
\def\showauthors@on{T}

\DeclareCaptionStyle{ruled}{labelfont=normalfont,labelsep=colon,strut=off} 
\floatstyle{ruled}
\newfloat{listing}{tb}{lst}{}
\floatname{listing}{Listing}
\def\showauthors@on{T}
\title{Evolution Meets Diffusion: Efficient Neural Architecture Generation}
\author{
    Bingye Zhou\textsuperscript{\rm 1}, Caiyang Yu\textsuperscript{\rm 1}, Chenwei Tang\textsuperscript{\rm 1}
}
\affiliations{
    \textsuperscript{\rm 1}Sichuan University\\

    xxc8865@gmail.com, jerome0324@163.com, tangchenwei@scu.edu.cn
}

\def\showauthors@on{T}
\begin{document}

\maketitle

\begin{abstract}
Neural Architecture Search (NAS) has gained widespread attention for its transformative potential in deep learning model design. 
However, the vast and complex search space of NAS leads to significant computational and time costs. 
Neural Architecture Generation (NAG) addresses this by reframing NAS as a generation problem, enabling the precise generation of optimal architectures for specific tasks. 
Despite its promise, mainstream methods like diffusion models face limitations in global search capabilities and are still hindered by high computational and time demands.
To overcome these challenges, we propose \underline{\textbf{E}}volutionary \underline{\textbf{D}}iffusion-based \underline{\textbf{N}}eural \underline{\textbf{A}}rchitecture \underline{\textbf{G}}eneration (EDNAG), a novel approach that achieves efficient and network-free architecture generation. 
EDNAG leverages evolutionary algorithms to mimic the iterative refinement process of diffusion models, using fitness to guide the evolution of candidate architectures from diverse random populations to task-optimal solutions. Through this evolution-driven refinement process, our method achieves end-to-end acceleration of architecture generation, delivering optimal solutions in a fraction of the time required by conventional network-involved diffusion-based approaches.
Extensive experiments demonstrate that EDNAG achieves SOTA performance in architecture optimization, with an improvement in accuracy of up to 10.45\%. 
Furthermore, it eliminates the need for time-consuming training and boosts inference speed by an average of 50$\times$, showcasing its exceptional efficiency. 
\end{abstract}


\section{Introduction}
\begin{figure}[tb]
    \centering
    \includegraphics[width=\linewidth]{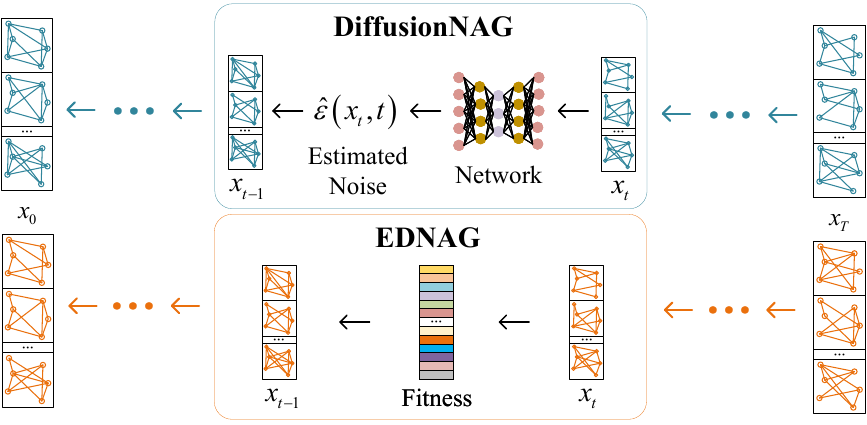}
    \caption{Overview of EDNAG. Unlike DiffusionNAG, which uses a trained noise prediction network to guide the denoising process, EDNAG employs a fitness-guided denoising strategy to derive the samples of next generation.}
    \label{introduction_figure}
\end{figure}

Neural Architecture Search (NAS)~\cite{dong2024automated,kang2023neural} represents an effective approach for the automated design of deep neural network architectures. 
However, conventional NAS methods are limited by the significant computational costs associated with exploring vast and intricate search spaces. 
Recently, the paradigm of Neural Architecture Generation (NAG)~\cite{nasir2024llmatic,lukasik2022learning} has revolutionized this optimization challenge by reframing it from an optimization task into a generation task, thus mitigating the limitations imposed by large search spaces. 
The NAG framework leverages advanced generation techniques to directly produce optimized neural architectures that are finely tuned for specific downstream applications, offering more scalable and resource-efficient solutions.

Previous architecture generation techniques can be categorized into two primary categories: Variational Auto-Encoders (VAE)-based~\cite{lin2023movae} and Generative Adversarial Networks (GAN)-based~\cite{huang2023scggan} methods. 
VAE-based methods train an encoder to map architecture features into embedding representations and a decoder to generate them from the latent space. In contrast, GAN-based methods employ a generator to produce architectures that receive feedback from a discriminator. 
However, both VAE-based and GAN-based methods require the simultaneous training of two networks, leading to significant challenges in high computational complexity, training stability~\cite{brock2018large}, and generation quality. 
Recently, diffusion-based methods have demonstrated superior performance by training a single network and generating architectures with multiple denoising steps, thereby improving stability and generation quality. 
For example, DiffusionNAG~\cite{an2023diffusionnag} and DiNAS~\cite{asthana2024multi} employ a graph diffusion framework, wherein the denoising process is iteratively conducted by a neural network trained to predict and remove noise. 
However, the need for thousands of such iterations results in significant computational overhead and slow generation speed.
Therefore, diffusion-based methods still face two major challenges: \textbf{(1) Time-consuming network training}. Training UNet or Score-Net demands substantial computational resources, and retraining is always required when transitioning to different search spaces; and \textbf{(2) Local optimality.} Naive diffusion models frequently converge to local optima, resulting in the generation of suboptimal architectures, especially when confronted with complex and extensive search spaces. 

To the end, we propose a novel and efficient \underline{\textbf{E}}volutionary \underline{\textbf{D}}iffusion-based \underline{\textbf{N}}eural \underline{\textbf{A}}rchitecture \underline{\textbf{G}}eneration Framework, called \textbf{EDNAG}.
The core of EDNAG is to achieve a more globally optimal and efficient neural architecture generation.

we replace the traditional multi-step network-involved denoising process by introducing evolutionary algorithms to simulate the denoising process in diffusion models.
This enables us to generate and refine architectures more efficiently without relying on repeated network inference, thereby avoiding time-consuming network training and significantly accelerating the generation process, as shown in \Cref{introduction_figure}.
Specifically, we propose a Fitness-guided Denoising (FD) strategy to produce a new generation of architectures from the previous ones in each denoising iteration, which prioritizes the exploration of high-potential regions.
Through multiple denoising iterations incorporating selection strategies and random mutations, the architecture samples progressively evolve towards high-fitness regions while simultaneously avoiding local optima, thereby generating high-performing and globally optimalal neural architectures.


We perform extensive experiments within NAS-Bench-201~\cite{dong2020bench} search space, demonstrating that EDNAG not only achieves SOTA performance in architecture generation with up to 10.45\% improvement in accuracy, but also exhibits an average 50$\times$ speedup compared to previous baselines. 
Furthermore, we expand the search spaces to TransNASBench-101~\cite{duan2021transnas}, MobileNetV3~\cite{howard2019searching}, DARTS~\cite{liu2018darts} and AutoFormer~\cite{chen2021autoformer}, demonstrating EDNAG's adaptability to extensive search spaces and its transferability to diverse downstream tasks. 

In summary, our contributions are as follows.
\begin{itemize}
    \item We propose a novel Evolutionary Diffusion-based framework for Neural Architecture Generation, called EDNAG, which simulates the denoising process in diffusion models within an evolutionary paradigm. EDNAG combines the superior generation quality of diffusion models with the global optimization capabilities of evolutionary algorithms. 
    \item We introduce a Fitness-guided Denoising (FD) strategy, which utilizes fitness instead of trained networks to progressively generate offspring architecture samples from previous ones, achieving the first network-free neural architecture generation approach. 
    \item Extensive experiments have demonstrated the SOTA performance of EDNAG in architecture generation with significantly fewer computational resources, as well as its outstanding adaptability and transferability. 
\end{itemize}


\section{Related Work}
\subsection{Neural Architecture Generation}
Neural architecture generation~\cite{dong2024automated,kang2023neural} has evolved from VAE-based~\cite{lin2023movae} to GAN-based~\cite{huang2023scggan} methods and has recently adopted diffusion-based~\cite{an2023diffusionnag} techniques. 
NAVIGATOR-D3~\cite{hemmi2024navigator} is a VAE-based NAG method that extracts the graph features of architectures by an encoder and then reconstructs them through a decoder to generate optimal neural architectures. 
Following the paradigm of GAN, GA-NAS~\cite{such2020generative} trains a discriminator to distinguish winning architectures and a generator to produce architectures based on the rewards provided by the discriminator, constituting an adversarial optimization problem. 
DiffusionNAG~\cite{an2023diffusionnag} represents the groundbreaking application of diffusion models in NAG, producing high-performing architectures through a denoising process with a trained score network. 
Rather than a continuous graph diffusion framework as in DiffusionNAG, DiNAS~\cite{asthana2024multi} leverages a discrete graph diffusion model with a score network to generate neural architectures. 
However, both DiffusionNAG and DiNAS require numerous denoising iterations to generate architecture samples, along with frequent network inferences during iterations, leading to inefficiency in the generation process. 
Their iterative denoising process also suffers from slow convergence and a tendency to stagnate in local optima. 
Moreover, the aforementioned methods require the training process of one or more networks for generation tasks, which consume significant computational resources, while our method, EDNAG, uniquely achieves a network-free and training-free approach. 

\subsection{Evolutionary Algorithms}
Evolutionary algorithms~\cite{slowik2020evolutionary,vikhar2016evolutionary}, designed for optimization problems, serve as population-based heuristic paradigms simulating the natural evolution of species, which are appreciated for their global optimization capabilities. 
Recent studies have revealed potential connections between generative models and evolutionary algorithms, and have begun preliminary attempts to address challenging optimization problems with diffusion models. 
\cite{zhang2024diffusion} mathematically demonstrates the equivalence between diffusion and evolution, conceptualizing the diffusion process as reversed evolution and the denoising process as evolution. 
\cite{hartl2024heuristically} improves evolutionary algorithms by integrating diffusion models, enabling the retention of historical information across generations. 
Indeed, the aforementioned studies have largely focused on leveraging the properties of diffusion models to enhance evolutionary algorithms for optimization tasks. 
However, no prior work has attempted to integrate the principles of evolutionary algorithms into diffusion models for conditional generation tasks. 
Meanwhile, neural architecture generation, as a conditional generation task aimed at producing optimal neural architectures, naturally highlights the potential of integrating evolutionary algorithms into NAG, forming a foundational motivation for EDNAG. 


\section{Method}  

\subsection{Preliminary}

Denoising Diffusion Implicit Model (DDIM)~\cite{song2020denoising} is a representative diffusion model. 
It involves a forward diffusion process, adding Gaussian noise to the original data, and a reverse denoising process, where the model learns to generate the original data by reversing the diffusion process. 

In the forward process, we blend data with Gaussian noise, perturbing the original data $x_0$ to the final diffused data $x_T$ as follows: 
\begin{equation}
\label{forward_diffusion}
x_t = \sqrt{\overline{\alpha_t}} x_0 + \sqrt{1 - \overline{\alpha_t}} \epsilon_t, \enspace \epsilon_t \sim \mathcal{N}(0, I),
\end{equation}
where $\overline{\alpha}_t = \prod_{i=1}^{T} \alpha_i$, representing the portion of data that remains undisturbed. 
In the reverse process, DDIM removes the noise from the randomly initialized data step by step, which is represented as: 
\begin{equation}
\begin{aligned}
    &x_{t-1}=\sqrt{\overline{\alpha_{t-1}}} \underbrace{\left(\frac{x_{t}-\sqrt{1-\overline{\alpha_{t}}} \hat{\epsilon}_{\theta}\left(x_{t}, t\right)}{\sqrt{\overline{\alpha_{t}}}}\right)}_{\text {predicted } x_{0} }+\\
    &\underbrace{\sqrt{1-\overline{\alpha_{t-1}}-\sigma_{t}^{2}} \cdot \hat{\epsilon}_{\theta}\left(x_{t}, t\right)}_{\text {direction pointing to } x_{t}}+\underbrace{\sigma_{t} \epsilon_{t}}_{\text {random noise}}.
    \label{ddim_denoise}
\end{aligned}
\end{equation} 
where $\sigma_t$ controls the degree of random perturbations $\epsilon_t \sim \mathcal{N}(0, I)$ during the denoising process. 
Notably, DDIM leverages a neural network $\epsilon_\theta$ to predict perturbed noise $\hat{\epsilon}_{\theta}\left(x_{t}, t\right)$ at time step $t$, guiding the denoising process. 
By iteratively repeating this denoising process in \Cref{ddim_denoise}, DDIM progressively generates the final samples. 
Our method, EDNAG, performs architecture generation based on the denoising process of DDIM, but eliminates the need for a network to estimate $\hat{\epsilon}_{\theta}\left(x_{t}, t\right)$.


\subsection{Evolutionary Diffusion-based NAG}
\label{evodiff_section}

\subsubsection{Overview}
\label{evodiff_section_overview}
\begin{figure}[tb]
    \centering
    \includegraphics[width=\linewidth]{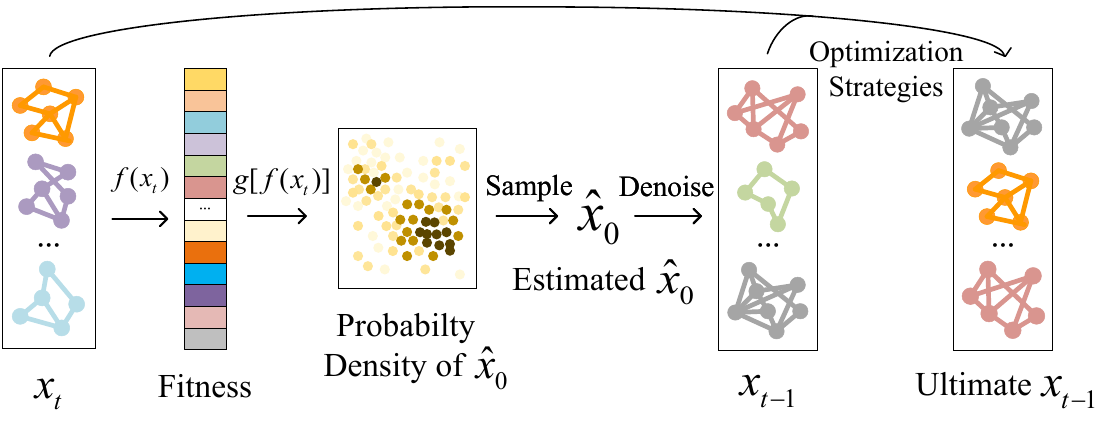}
    \caption{Fitness-guided denoising process. EDNAG generates neural architectures through iterative denoising process with FD strategy. In each iteration, FD strategy maps the fitness of $x_t$ to the probability density of optimal samples $x_0$, subsequently uses the DDIM denoising formula to obtain $x_{t-1}$, and finally employs optimization strategies to select the final samples $x_{t-1}$ from $x_t$ and denoised $x_{t-1}$.}
    \label{denoise_method_figure}
\end{figure}

EDNAG generates neural architectures through an iterative denoising process. 
In each iteration, EDNAG first evaluates the fitness of the architecture population using a neural predictor, then employs the fitness-guided denoising (FD) strategy to generate offspring samples, and finally leverages optimization strategies to select architecture samples. 
The pipeline of EDNAG is described in ~\Cref{denoise_method_figure}, and the pseudocode is displayed in ~\Cref{generation_algorithm}. 

\begin{algorithm}[tb]
    \caption{Neural Architecture Generation Process}
    \label{generation_algorithm}
    \textbf{Input}: Population size $N$, number of nodes $D_1$, types of operations $D_2$, denoising steps $T$, fitness evaluator $f$, mapping function $g$.\\
    \textbf{Output}: Generated optimal architectures $x_0$.
    
    \begin{algorithmic}[1] 
        \STATE Sample $x_T \sim \mathcal{N}(0, I^{N \times D_1 \times D_2})$.
        \FOR{$t$ in $[T,T-1,\ldots,1]$}
        \STATE Map fitness $f(x_t)$ to probability density $g[f(x_t)]$.
        \STATE Denoise from $x_t$ to $\hat{x}_{t-1}$ according to $g[f(x_t)]$.
        \STATE Select final samples $x_{t-1}$ at the $t-1$-th time step from parent $x_t$ and denoised offspring $\hat{x}_{t-1}$ by optimization strategies.
        \ENDFOR
        \STATE \textbf{return} $x_0$
    \end{algorithmic}
\end{algorithm}

In the following sections, we first explain the fitness-guided denoising (FD) strategy. 
Next, we describe our optimization strategies for selecting architecture samples. 
More detailed derivations and explanations are provided in the Appendix. 

\subsubsection{Denoising Process of EDNAG}
\label{evodiff_section_denoising}
FD strategy simulates the denoising process of DDIM to derive new samples $x_{t-1}$ from previous samples $x_t$, generating offspring samples and driving the evolutionary process towards optimal solutions. 
By iteratively performing $T$ denoising steps with FD strategy, the initial architecture samples $x_T$ (random noise samples) can be progressively transformed into optimal architecture samples $x_0$. 

For each denoising iteration, let $\hat{x}_0$ denotes the predicted $x_0$ in \Cref{ddim_denoise}:  
\begin{equation}
\label{replace_x0_formula}
    \hat{x}_0 = \frac{x_{t}-\sqrt{1-\overline{\alpha_{t}}} \hat{\epsilon}_{\theta}\left(x_{t}, t\right)}{\sqrt{\overline{\alpha_{t}}}}.
\end{equation}
With \Cref{replace_x0_formula}, the denoising process can be transformed into \Cref{main_denoise}.
\begin{equation}
\begin{aligned}
\label{main_denoise}
    x_{t-1} &= \sqrt{\overline{\alpha_{t-1}}} \hat{x}_0 + \sigma_{t} \epsilon_{t}\\
    + & \sqrt{1-\overline{\alpha_{t-1}}-\sigma_{t}^{2}} \cdot \frac{x_{t}-\sqrt{\overline{\alpha_{t}}}\hat{x}_0}{\sqrt{1-\overline{\alpha_{t}}}}.\\
\end{aligned}
\end{equation}
Therefore, to perform the denoising process from $x_t$ to $x_{t-1}$, it is only necessary to determine $\hat{x}_0$ (estimated optimal architectures). 

\begin{figure}[tb]
    \centering
    \includegraphics[width=\linewidth]{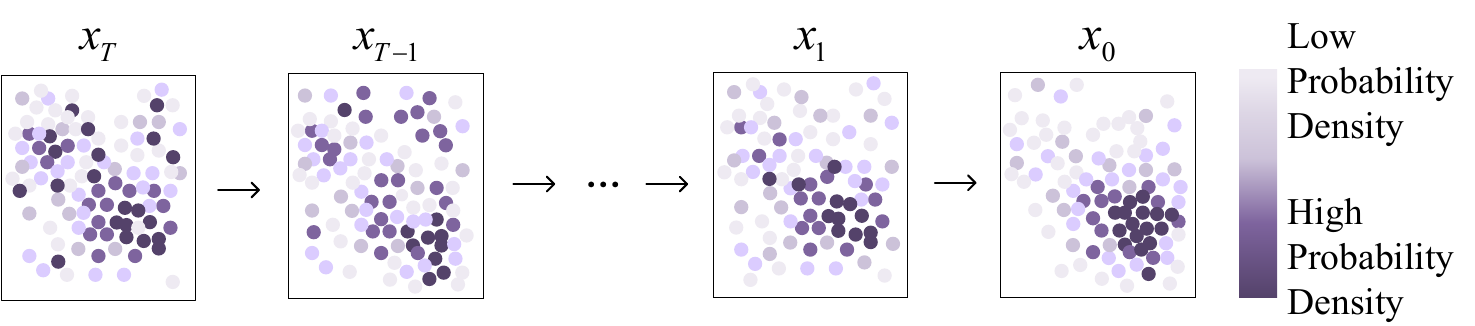}
    \caption{Denoising process of EDNAG. Individuals with higher fitness are more likely to be preserved throughout the denoising process, leading to an increased probability density in the final sample distribution.}
    \label{iteratiions}
\end{figure}

According to the Bayes’ theorem, the optimal architectures $\hat{x}_0$ can be estimated from the samples $x_t$ at time step $t$ and their fitness $f(x_t)$, as detailed below: 
\begin{equation}
\label{derive_of_x0}
    p(\hat{x}_0=x|x_t)=\frac{p(\hat{x}_0=x)\cdot p(x_t|x_0=x)}{p(x_t)},
\end{equation}
where $p(x_t|x_0=x)$ is derived from the forward diffusion process described by \Cref{forward_diffusion}, as shown below: 
\begin{equation}
\label{forward_possibility}
    p(x_t|x_0=x)\sim \mathcal{N}(x_t; \sqrt{\overline{\alpha}_t}x, 1 - \overline{\alpha}_t).
\end{equation}
In order to get $p(\hat{x}_0=x)$, we should estimate $\hat{x}_0$ by mapping the distribution of $x_t$ to the distribution of $\hat{x}_0$ with fitness guidance. 
Specifically, we view the denoising process of DDIM as the evolutionary process of GAs, both transforming samples from an initial random Gaussian distribution to an ultimate optimal distribution progressively~\cite{zhang2024diffusion}. 
We consider that individuals with higher fitness are more likely to be retained during evolution, resulting in a higher probability of appearing in the final samples $\hat{x}_0$, as shown in \Cref{iteratiions}. 
Therefore, a higher fitness corresponds to a higher probability density in $\hat{x}_0$. 
We model this relationship as a mapping function $g(x)$, which maps the fitness $f(x_t)$ of $x_t$ to the probability density $p(\hat{x}_0)$ of $\hat{x}_0$:
\begin{equation}
    p(\hat{x}_0=x) = g[f(x)].
\end{equation}

We further define $x_t$ as the previous samples consisting of $N$ architectures, where $x_t = [\boldsymbol{x}_t^1, \boldsymbol{x}_t^2, \ldots, \boldsymbol{x}_t^N]$. Combining \Cref{derive_of_x0,forward_possibility}, $\hat{x}_0$ can be estimated as follows:
\begin{equation}
\begin{aligned}
\label{main_estimation_x_0}
    &\hat{x}_0 = \frac{1}{p(x_t)} \sum_{\boldsymbol{x}_t^i \in x_t}{g[f(\boldsymbol{x}_t^i)] \mathcal{N}(x_t; \sqrt{\overline{\alpha_t}} \boldsymbol{x}_t^i, 1 - \overline{\alpha_t}) \boldsymbol{x}_t^i},\\
    &p(x_t) = \sum_{\boldsymbol{x}_t^i \in x_t}{g[f(\boldsymbol{x}_t^i)] \mathcal{N}(x_t; \sqrt{\overline{\alpha_t}} \boldsymbol{x}_t^i, 1 - \overline{\alpha_t})},
\end{aligned}
\end{equation}
where $p(x_t)$ actually serves as a regularization term. 
In fact, $\hat{x}_0$ can be regarded as the weighted summation of each neural architecture $\boldsymbol{x}_t^i$ in samples $x_t$, where architectures with higher fitness have a greater influence on $\hat{x}_0$. 

Finally, with \Cref{main_denoise,main_estimation_x_0}, we can denoise from $x_t$ to $x_{t-1}$ as follows:
\begin{equation}
\left\{
\begin{aligned}
\label{final_total_denoise}
    \hspace{0.06em}&x_{t-1} = \sqrt{\overline{\alpha_{t-1}}} \hat{x}_0 + \sigma_{t} \epsilon_{t} \\
    &\quad\quad\quad + \sqrt{1-\overline{\alpha_{t-1}}-\sigma_{t}^{2}} \cdot \frac{x_{t}-\sqrt{\overline{\alpha_{t}}}\hat{x}_0}{\sqrt{1-\overline{\alpha_{t}}}},\\ 
    &\hat{x}_0\!=\!\frac{1}{p(x_t)}\!\sum_{\boldsymbol{x}_t^i \in x_t}\!{g[f(\boldsymbol{x}_t^i)] \mathcal{N}(x_t; \sqrt{\overline{\alpha_t}} \boldsymbol{x}_t^i, 1 - \overline{\alpha_t}) \boldsymbol{x}_t^i},\\
    &p(x_t) = \sum_{\boldsymbol{x}_t^i \in x_t}{g[f(\boldsymbol{x}_t^i)] \mathcal{N}(x_t; \sqrt{\overline{\alpha_t}} \boldsymbol{x}_t^i, 1 - \overline{\alpha_t})}.
\end{aligned}
\right.
\end{equation}
By iteratively applying the FD denoising strategy across $x_t$, $x_{t-1}$,..., $x_0$, which guides the evolution of architecture samples toward higher-fitness subspaces, EDNAG ultimately generates global-optimal neural architectures. 
In particular, $\epsilon_{t}$ represents stochastic noise, which serves as random mutations in GAs, thereby enabling the generation process to overcome local optima.

\subsubsection{Optimization Strategies for Selection}
\label{evodiff_section_selection}
In each denoising iteration, we apply \Cref{final_total_denoise} to denoise architecture samples from $x_t$ to $x_{t-1}$. 
Subsequently, we employ optimization strategies (denoted as $S$) to select the final $x'_{t-1}$ at time step $t-1$ from previous samples $x_t$ and denoised samples $x_{t-1}$:
\begin{equation}
    x'_{t-1} = S\left(x_t,x_{t-1}\right).
\end{equation}
Specifically, we adopt an elitist strategy to retain the fittest architectures, a diversity strategy to preserve diverse architectures, and finally the roulette wheel strategy to select remaining architectures.
During selection, 70\% of individuals are chosen by roulette wheel selection, 10\% by elitism, and the remaining 20\% by the diversity strategy that selects the most distinct individuals based on the Euclidean distance. 
These optimization strategies, which simulate natural selection in GAs, ensure task-optimality while simultaneously improving the stability of the generation process and promoting diversity within the generated architecture samples. 

\subsection{Transferable Neural Predictor for Guidance}
\label{predictor_section}

\begin{figure}[tb]
    \centering
    \includegraphics[width=\linewidth]{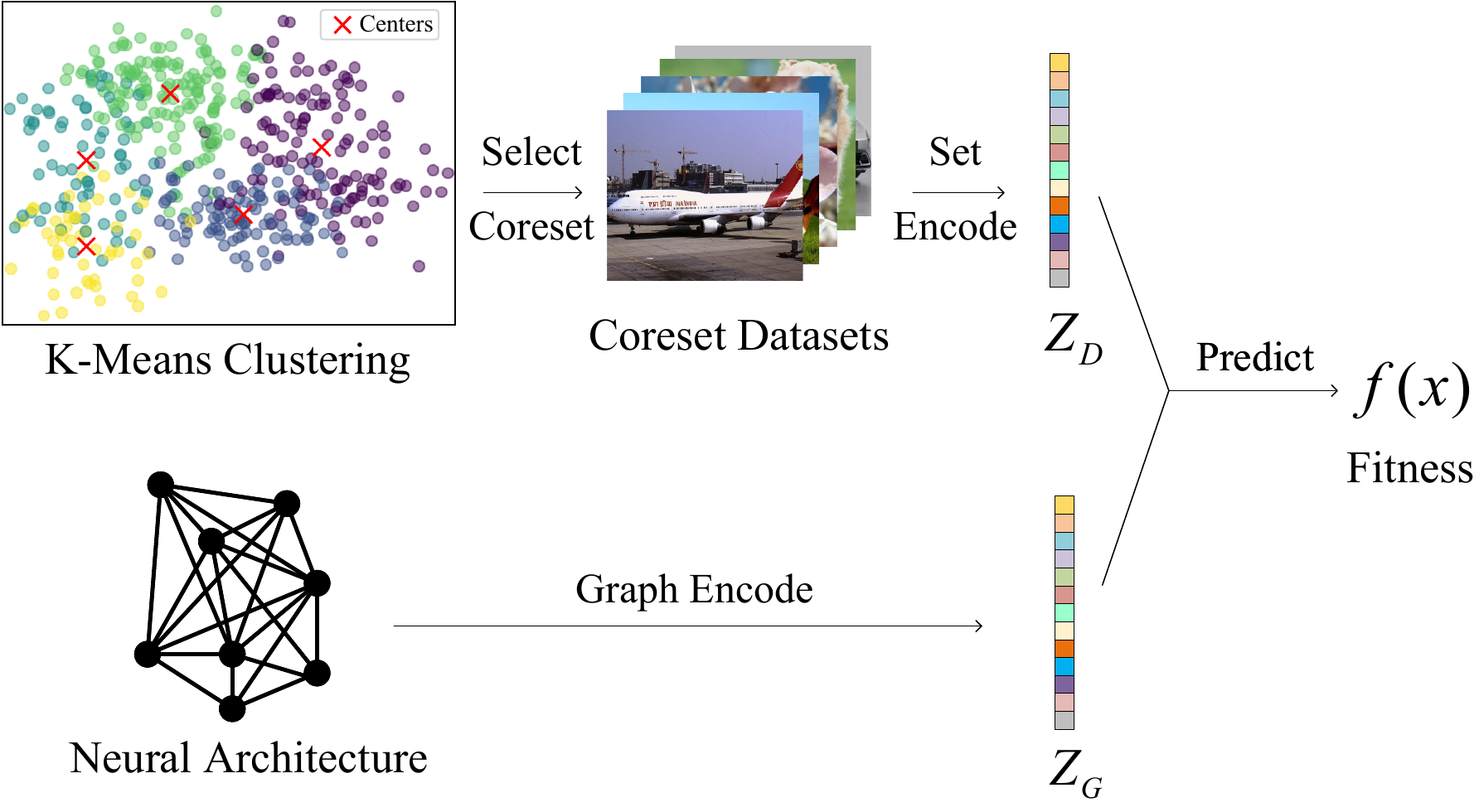}
    \caption{Neural predictor in EDNAG. During prediction, K-means clustering is applied to each class in $\mathcal{T}$, and the resulting centroids form the coreset $\mathcal{D}$. The set encoder $p(z_{\mathcal{D}} | \mathcal{D})$ and graph encoder $p(z_{\mathcal{G}} | \mathcal{G})$ embed $\mathcal{D}$ and neural architecture $\mathcal{G}$ into latent representations $z_{\mathcal{D}}$ and $z_{\mathcal{G}}$, respectively. The linear predictor $p(f(x) | z_{\mathcal{D}}, z_{\mathcal{G}})$ finally estimates the architecture’s performance.}
    
    \label{predictor_figure}
\end{figure}

For task-specific conditional neural architecture generation, fitness evaluation is critical in guiding the denoising process of generation, which is conducted by a transferable neural predictor. 
Following DiffusionNAG, our neural predictor comprises three components: a set encoder $p(z_{\mathcal{D}} | \mathcal{D})$\cite{lee2019set}, a graph encoder $p(z_{\mathcal{G}} | \mathcal{G})$\cite{zhang2019d}, and a linear predictor $p(f(x) | z_{\mathcal{D}}, z_{\mathcal{G}})$. 

In DiffusionNAG, the set encoder $p(z_{\mathcal{D}} | \mathcal{D})$ encodes a small subset of images randomly sampled from the task dataset for each prediction, which overlooks the uniformity of dataset distribution, resulting in biased and unstable predictions. 
To address these issues, we first select a coreset~\cite{liu2023dream} from the task dataset and then feed it into the set encoder $p(z_{\mathcal{D}} | \mathcal{D})$. 
Specifically, during prediction, each class of images in the task dataset $\mathcal{T} = {\left\{ \left( x^i, y^i \right) \right\}}^{|\mathcal{T}|}$ undergoes K-means clustering to select centroids of clusters as coreset $\mathcal{D} = {\left\{ \left( x^i, y^i \right) \right\}}^{|\mathcal{D}|}$, where $|\mathcal{D}|$ represents the number of samples per class and $|\mathcal{D}| \ll |\mathcal{T}|$. 
The entire process is detailed in \Cref{predictor_figure}. 

    

Notably, similar to previous methods, our neural predictor requires training. 
Although EDNAG shares this commonality with other NAG approaches based on generative models with proxy predictors, its key distinction and advantage lie in the generative model, which operates without score networks, thus significantly enhancing efficiency. 


\section{Experiments}
We conduct sufficient experiments in the NAS-Bench-201, TransNASBench-101, MobileNetV3, DARTS, and AutoFormer search spaces. 
Subsequently, we perform ablation experiments on the FD strategy and further experiments on generation efficiency. 
For experiments in NAS-Bench-101 and NAS-Bench-301, baseline descriptions, hyperparameter settings, sensitivity analysis of parameters, comprehensive comparisons of GPU time across 14 baselines, and more in-depth experiments, please refer to the Appendix. 

\subsection{Experiments on Neural Architecture Generation}
\label{main_exp}
\subsubsection{Experiment Setup}
We compare EDNAG with mainstream NAS methods, demonstrating its superior capabilities in architecture generation across various datasets. 
\textbf{One-shot NAS} performs NAS within sub-networks of a supernet, including weight inheritance-based and gradient-based methods. 
\textbf{BO-based NAS} employs Bayesian optimization techniques (e.g. Gaussian processes) to search for optimal neural architectures. 
\textbf{NAG} employs different generative models to directly generate neural architectures. 

For the generation process, the denoising step of EDNAG is $100$. The noise scale of diffusion ranges from $\alpha_T=10^{-4}$ to $\alpha_0=1-10^{-4}$ and the noise scale of denoising is $\sigma=0.8$. 
The population size (or batch size) is $100$ for MobileNetV3 search space and $30$ for other search spaces. 

\subsubsection{Experiments on NAS-Bench-201 Search Space}
\label{nb201_experiment}

\begin{table*}[tb]
    \centering
    {\fontsize{9}{11}\selectfont
        \def\arraystretch{1.1}
            \begin{tabular}{ccccccccccc}
                \hline\hline
                \rowcolor{tableheadcolor} 
                 &  & \multicolumn{2}{c}{CIFAR-10} & \multicolumn{2}{c}{CIFAR-100} & \multicolumn{2}{c}{Aircraft} & \multicolumn{2}{c}{Pets} \\
                \cline{3-10}
                \rowcolor{tableheadcolor} 
                \multirow{-2}{*}{Type} & 
                \multirow{-2}{*}{Method} & 
                Acc. (\%) & Archs & Acc. (\%) & Archs & Acc. (\%) & Archs & Acc. (\%) & Archs \\
                \hline\hline
                \multirow{4}{*}{One-shot} 
                    & \cellcolor{lightgray}RSPS~\shortcite{li2020random} & \cellcolor{lightgray}84.07$\pm$3.61 & \cellcolor{lightgray}N/A & \cellcolor{lightgray}52.31$\pm$5.77 & \cellcolor{lightgray}N/A & \cellcolor{lightgray}42.19$\pm$3.88 & \cellcolor{lightgray}N/A & \cellcolor{lightgray}22.91$\pm$1.65 & \cellcolor{lightgray}N/A \\
                    & SETN~\shortcite{dong2019one} & 87.64$\pm$0.00 & N/A & 59.09$\pm$0.24 & N/A & 44.84$\pm$3.96 & N/A & 25.17$\pm$1.68 & N/A \\
                    & \cellcolor{lightgray}PC-DARTS~\shortcite{xu2020pc} & \cellcolor{lightgray}93.66$\pm$0.17 & \cellcolor{lightgray}N/A & \cellcolor{lightgray}66.64$\pm$2.34 & \cellcolor{lightgray}N/A & \cellcolor{lightgray}26.33$\pm$3.40 & \cellcolor{lightgray}N/A & \cellcolor{lightgray}25.31$\pm$1.38 & \cellcolor{lightgray}N/A \\
                    & DrNAS~\shortcite{chen2021drnas} & 94.36$\pm$0.00 & N/A & \textbf{73.51$\pm$0.00} & N/A & 46.08$\pm$7.00 & N/A & 26.73$\pm$2.61 & N/A \\
                \hline
                \multirow{4}{*}{BO} 
                    & \cellcolor{lightgray}BANANAS~\shortcite{white2019bananas} & \cellcolor{lightgray}\textbf{94.37$\pm$0.00} & \cellcolor{lightgray}46 & \cellcolor{lightgray}\textbf{73.51$\pm$0.00} & \cellcolor{lightgray}88 & \cellcolor{lightgray}41.72$\pm$0.00 & \cellcolor{lightgray}40 & \cellcolor{lightgray}40.15$\pm$1.59 & \cellcolor{lightgray}17 \\
                    & NASBOWL~\shortcite{ru2020interpretable} & 94.34$\pm$0.00 & 100 & \textbf{73.51$\pm$0.00} & 87 & 53.73$\pm$0.83 & 40 & 41.29$\pm$1.10 & 17 \\
                    & \cellcolor{lightgray}HEBO~\shortcite{cowen2020hebo} & \cellcolor{lightgray}94.34$\pm$0.00 & \cellcolor{lightgray}100 & \cellcolor{lightgray}72.62$\pm$0.20 & \cellcolor{lightgray}100 & \cellcolor{lightgray}49.32$\pm$6.10 & \cellcolor{lightgray}40 & \cellcolor{lightgray}40.55$\pm$1.15 & \cellcolor{lightgray}18 \\
                    & TNAS~\shortcite{shala2023transfer} & \textbf{94.37$\pm$0.00} & 29 & \textbf{73.51$\pm$0.00} & 59 & \textbf{59.15$\pm$0.58} & 26 & 40.00$\pm$0.00 & 6 \\
                \hline
                \multirow{4}{*}{NAG}
                    & \cellcolor{lightgray}DiNAS~\shortcite{asthana2024multi} & \cellcolor{lightgray}\textbf{94.37$\pm$0.00} & \cellcolor{lightgray}192 & \cellcolor{lightgray}\textbf{73.51$\pm$0.00} & \cellcolor{lightgray}192  & \cellcolor{lightgray}- & \cellcolor{lightgray}- & \cellcolor{lightgray}- & \cellcolor{lightgray}- \\
                    & MetaD2A~\shortcite{lee2021rapid} & \textbf{94.37$\pm$0.00} & 100 & 73.34$\pm$0.04 & 100 & 57.71$\pm$0.20 & 40 & 39.04$\pm$0.20 & 40 \\
                    & \cellcolor{lightgray}DiffusionNAG~\shortcite{an2023diffusionnag} & \cellcolor{lightgray}\textbf{94.37$\pm$0.00} & \cellcolor{lightgray}\underline{1} & \cellcolor{lightgray}\textbf{73.51$\pm$0.00} & \cellcolor{lightgray}2 & \cellcolor{lightgray}58.83$\pm$3.75 & \cellcolor{lightgray}3 & \cellcolor{lightgray}41.80$\pm$3.82 & \cellcolor{lightgray}2 \\
                    & \textbf{EDNAG} (Ours) & \textbf{94.37$\pm$0.00} & \textbf{1} & \textbf{73.51$\pm$0.00} & \textbf{1} & \textbf{60.14$\pm$0.89} & \textbf{1} & \textbf{46.17$\pm$0.67} & \textbf{1} \\
                \hline
                \rowcolor{lightgray} 
                \multicolumn{1}{c}{\cellcolor{lightgray}} & Global Best & 94.37 & N/A & 73.51 & N/A & - & N/A & - & N/A \\
                \hline\hline
            \end{tabular}
        \caption{Experiments on NAS-BENCH-201 search space. Acc refers to the average accuracy over 3 runs with 95\% confidence intervals, and Archs denotes the number of neural architectures actually trained to achieve this accuracy.}
        \label{nb201_experiments}
    }
\end{table*}

NAS-BENCH-201 is a widely used cell-based search space comprising 15,625 distinct neural architectures.
Within NAS-BENCH-201, we perform conditional generation of architectures across four task datasets and subsequently train the top-performing architecture to validate their performance, as shown in \Cref{nb201_experiments}. 
Across diverse task datasets, EDNAG outperforms all previous methods, achieving SOTA performance and showcasing exceptional capability in generating optimal architectures. 
This is attributed to the FD strategy, which efficiently directs architecture sampling toward high-fitness subspaces, effectively steering the generation process toward global optima.
Compared to DiffusionNAG, the previous SOTA baseline directly using naive diffusion models with limited optimization abilities, EDNAG excels with its superior global optimization capabilities while maintaining efficient generative abilities. 

Moreover, EDNAG also significantly improves the efficiency of architecture validation by reducing the number of architectures (Archs) requiring training to achieve final accuracy. 
In all datasets, the final accuracies are obtained from the NAS-Bench-201 benchmarks~\cite{dong2020bench} by querying only the \textbf{top-1} generated architecture. 
The results illustrate that the top-1 neural architecture identified by our neural predictor, attains optimal accuracy after real training, achieved by the application of coreset selection before dataset encoding. 
Unlike the random sampling method used in DiffusionNAG, our approach significantly enhances the reliability of the predicted accuracies.

\subsubsection{Experiments on TransNASBench-101 Search Space}
\label{trans101_experiment}

\begin{table*}[tb]
    \centering
    {\fontsize{9}{11}\selectfont
        \def\arraystretch{1.1}
            \begin{tabular}{ccccccccc}
                \hline\hline
                \rowcolor{tableheadcolor} 
                 &  & OC & SC & AE & SE & SS & RP & JS \\
                \cline{3-9}
                \rowcolor{tableheadcolor} 
                \multirow{-2}{*}{Level} & 
                \multirow{-2}{*}{Method} & 
                Acc. (\%) & Acc. (\%) & SSIM & SSIM & mIoU & L2 Loss & Acc. (\%) \\
                \hline\hline
                \multirow{5}{*}{Micro}
                    & \cellcolor{lightgray}ZiCo~\shortcite{li2024theoretically} & \cellcolor{lightgray}44.72$\pm$0.6  & \cellcolor{lightgray}53.70$\pm$0.4 & \cellcolor{lightgray}50.01$\pm$0.2 & \cellcolor{lightgray}57.50$\pm$0.5 & \cellcolor{lightgray}25.11$\pm$0.5 & \cellcolor{lightgray}61.57$\pm$0.6 & \cellcolor{lightgray}93.20$\pm$0.4 \\
                    & GEA~\shortcite{LOPES2024127509} & 45.98$\pm$0.2 & 54.85$\pm$0.1 & 57.11$\pm$0.3 & 58.33$\pm$1.0 & 25.95$\pm$0.2 & 59.93$\pm$0.5 & 94.96$\pm$0.2 \\
                    & \cellcolor{lightgray}CATCH~\shortcite{chen2020catch} & \cellcolor{lightgray}45.27$\pm$0.5 & \cellcolor{lightgray}54.38$\pm$0.2 & \cellcolor{lightgray}56.13$\pm$0.7 & \cellcolor{lightgray}56.99$\pm$0.6 & \cellcolor{lightgray}25.38$\pm$0.4 & \cellcolor{lightgray}60.70$\pm$0.7 & \cellcolor{lightgray}94.25$\pm$0.3 \\
                    & \textbf{EDNAG} (Ours) & \textbf{46.28$\pm$0.04} & \textbf{54.91$\pm$0.03} & \textbf{57.63$\pm$0.09} & \textbf{59.26$\pm$0.36} & \textbf{26.19$\pm$0.08} & \textbf{59.01$\pm$0.30} & \textbf{95.19$\pm$0.18} \\ 
                \cline{2-9}
                    & \cellcolor{lightgray}Global Best & \cellcolor{lightgray}46.32 & \cellcolor{lightgray}54.94 & \cellcolor{lightgray}57.72 & \cellcolor{lightgray}59.62 & \cellcolor{lightgray}26.27 & \cellcolor{lightgray}58.71 & \cellcolor{lightgray}95.37 \\
                \hline
                \multirow{5}{*}{Macro}
                    & \cellcolor{lightgray}REA~\shortcite{real2019regularized} & \cellcolor{lightgray}47.09$\pm$0.4 & \cellcolor{lightgray}56.57$\pm$0.4 & \cellcolor{lightgray}69.98$\pm$3.6 & \cellcolor{lightgray}60.88$\pm$1.0 & \cellcolor{lightgray}28.87$\pm$0.4 & \cellcolor{lightgray}58.73$\pm$1.1 & \cellcolor{lightgray}96.88$\pm$0.1 \\
                    & PPO~\shortcite{schulman2017proximal} & 46.84$\pm$0.4 & 56.48$\pm$0.3 & 70.92$\pm$3.2 & 60.82$\pm$0.8 & 28.31$\pm$0.5 & 58.84$\pm$1.1 & 96.76$\pm$0.2 \\
                    & \cellcolor{lightgray}CATCH~\shortcite{chen2020catch} & \cellcolor{lightgray}47.29$\pm$0.3 & \cellcolor{lightgray}56.49$\pm$0.3 & \cellcolor{lightgray}70.36$\pm$3.0 & \cellcolor{lightgray}60.85$\pm$0.7 & \cellcolor{lightgray}28.71$\pm$0.4 & \cellcolor{lightgray}59.37$\pm$0.6 & \cellcolor{lightgray}96.82$\pm$0.1 \\
                    & \textbf{EDNAG} (Ours) & \textbf{47.69$\pm$0.27} & \textbf{57.29$\pm$0.19} & \textbf{75.90$\pm$0.98} & \textbf{63.50$\pm$0.85} & \textbf{29.60$\pm$0.06} & \textbf{56.20$\pm$0.52} & \textbf{96.98$\pm$0.04} \\ 
                \cline{2-9}
                    & \cellcolor{lightgray}Global Best & \cellcolor{lightgray}47.96 & \cellcolor{lightgray}57.48 & \cellcolor{lightgray}76.88 & \cellcolor{lightgray}64.35 & \cellcolor{lightgray}29.66 & \cellcolor{lightgray}55.68 & \cellcolor{lightgray}97.02 \\
                \hline\hline
            \end{tabular}
        \caption{Experiments on TransNASBench-101 across seven tasks. Metrics are reported with 95\% confidence intervals over 3 runs. For improved readability, the values in the table for SSIM, mIoU, and L2 Loss are scaled by 100.}
        \label{transnasbench101_experiments}
    }
\end{table*}

TransNASBench-101 extends downstream tasks beyond basic image classification, encompassing a diverse set of 7 tasks: Object Classification (OC), Scene Classification (SC), Autoencoding (AE), Surface Normal Estimation (SE), Semantic Segmentation (SS), Room Layout Prediction (RP), and Jigsaw Puzzle Solving (JS). 
Moreover, it defines two types of search spaces: a cell-based micro search space and a stack-based macro skeleton search space, which encompass 4,096 and 3,256 unique architectures, respectively. 

We conduct experiments in both micro-level and macro-level search spaces, as presented in \Cref{transnasbench101_experiments}. 
Across both search spaces, EDNAG demonstrates substantial performance enhancements of up to \textbf{7.02}\% over prior baselines, consistently approaching global optimality across all downstream tasks while maintaining stability across three independent runs. 
These findings not only demonstrate the effectiveness of the FD strategy in improving global optimization capabilities for generative models, but also highlight the transferability of EDNAG for cross-task neural architecture generation within various downstream scenarios. 

\subsubsection{Experiments on MobileNetV3 and DARTS  Search Spaces}
\label{larger_mbv3_experiment}

\begin{figure}[tb]
    \centering
    \includegraphics[width=\linewidth]{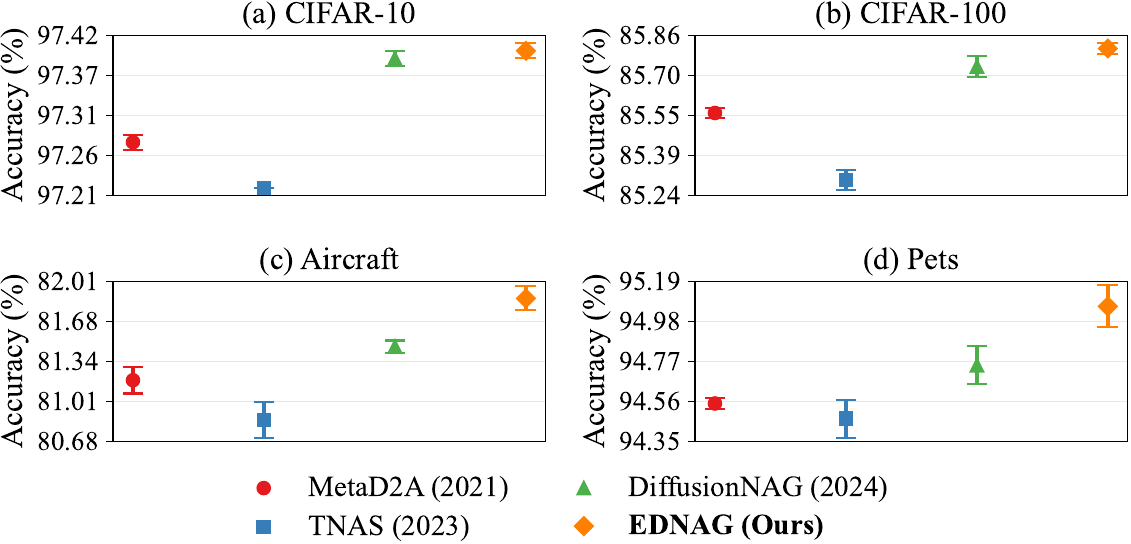}
    \caption{Experiments on the MobileNetV3 search space.}
    \label{mbv3_experiments}
\end{figure}

\begin{figure}[tb]
    \centering
    \includegraphics[width=\linewidth]{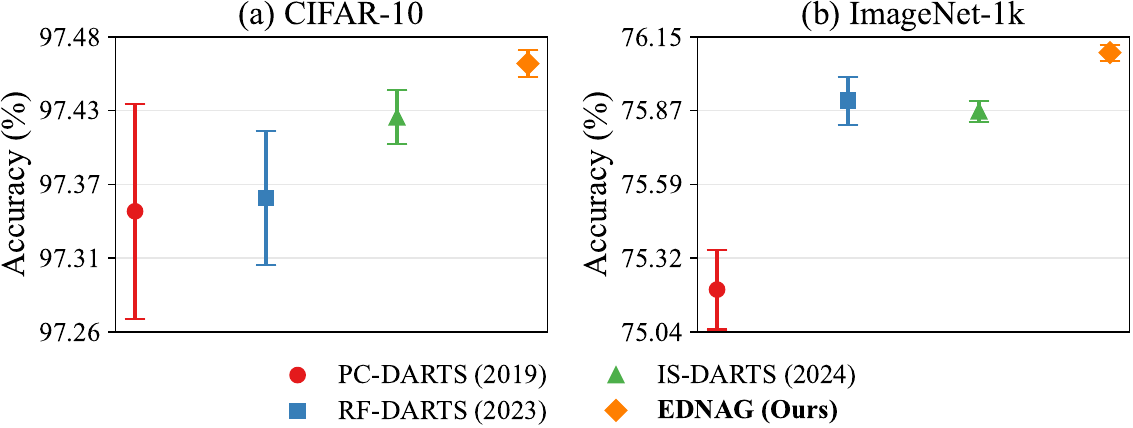}
    \caption{Experiments on the DARTS search space.}
    \label{darts_experiments}
\end{figure}

Due to the relatively limited scale of the above search spaces, our study further extends to more complex search spaces to demonstrate the scalability of EDNAG. 
MobileNetV3 is a large search space which encompasses over $10^{19}$ distinct neural architectures, and DARTS~\cite{liu2018darts} is a well-known search space consisting over $10^{18}$ architectures. 
In addition, we extend our task dataset to ImageNet-1k~\cite{deng2009imagenet}, a classic large-scale dataset which spans 1,000 categories and contains over $10^6$ images. 
In both search spaces, we report the average accuracy of different methods over 3 runs with with 95\% confidence intervals.

We conduct experiments on four distinct task datasets within MobileNetV3, as shown in \Cref{mbv3_experiments}, where EDNAG demonstrates an improvement in accuracy of up to \textbf{0.49}\%. 
We also conduct experiments on CIFAR-10 and ImageNet-1k within DARTS, as presented in \Cref{darts_experiments}, where EDNAG achieves an accuracy improvement of up to \textbf{0.24}\%.
This experiment not only demonstrates the superior ability of architecture generation for EDNAG but also confirms its adaptability and effectiveness in large-scale search spaces and task datasets. 

Moreover, when transitioning to different search spaces, DiffusionNAG necessitates retraining the score network within the diffusion model to accommodate the new dimensionality of matrix, which is troublesome and time-consuming.
In contrast, as a training-free NAG method, EDNAG adapts to new search spaces without training or retraining, incurring only negligible computational cost.

\subsubsection{Experiments on AutoFormer Search Space}
\label{autoformer_experiment}

\begin{table}[tb]
    \centering
    {\fontsize{9}{11}\selectfont
        \def\arraystretch{1.1}
        \begin{tabular}{ccccccc}
            \hline\hline
            \rowcolor{tableheadcolor} 
            & & \#Params & Top-1 & Top-5 \\
            \cline{3-5}
            \rowcolor{tableheadcolor} 
            \multirow{-2}{*}{Type} & 
            \multirow{-2}{*}{Method} & 
            (M) & (\%) & (\%) \\
            \hline\hline
            \multirow{4}{*}{Tiny} 
                & \cellcolor{lightgray}AutoFormer~\shortcite{chen2021autoformer} & \cellcolor{lightgray}\textbf{5.7} & \cellcolor{lightgray}74.7 & \cellcolor{lightgray}92.6 \\
                & TF-TAS-Ti~\shortcite{zhou2022training} & 5.9 & \textbf{75.3} & 92.8 \\
                & \cellcolor{lightgray}PiMO-NAS-A~\shortcite{luo2024pareto} & \cellcolor{lightgray}\textbf{5.7} & \cellcolor{lightgray}75.1 & \cellcolor{lightgray}92.7 \\
                & \textbf{EDNAG} (Ours) & \textbf{5.7} & \textbf{75.3} & \textbf{92.9} \\
            \hline
            \multirow{4}{*}{Small} 
                & \cellcolor{lightgray}AutoFormer~\shortcite{chen2021autoformer} & \cellcolor{lightgray}22.9 & \cellcolor{lightgray}81.7 & \cellcolor{lightgray}95.7 \\
                & TF-TAS-Ti~\shortcite{zhou2022training} & 22.8 & 81.9 & \textbf{95.8} \\
                & \cellcolor{lightgray}PiMO-NAS-A~\shortcite{luo2024pareto} & \cellcolor{lightgray}21.4 & \cellcolor{lightgray}81.6 & \cellcolor{lightgray}95.5 \\
                & \textbf{EDNAG} (Ours) & \textbf{21.2} & \textbf{82.0} & \textbf{95.8} \\
            \hline
            \multirow{4}{*}{Base} 
                & \cellcolor{lightgray}AutoFormer~\shortcite{chen2021autoformer} & \cellcolor{lightgray}54.0 & \cellcolor{lightgray}\textbf{82.4} & \cellcolor{lightgray}95.7 \\
                & TF-TAS-Ti~\shortcite{zhou2022training} & 54.0 & 82.2 & 95.6 \\
                & \cellcolor{lightgray}PiMO-NAS-A~\shortcite{luo2024pareto} & \cellcolor{lightgray}\textbf{51.7} & \cellcolor{lightgray}82.3 & \cellcolor{lightgray}95.7 \\
                & \textbf{EDNAG} (Ours) & 51.9 & \textbf{82.4} & \textbf{95.8} \\
            \hline\hline
        \end{tabular}
        \caption{Experiments on the AutoFormer search space. \#Parameters, Top-1 Accuracy and Top-5 Accuracy are reported with 3 repeated runs with classification task on ImageNet-1k with 95\% confidence.}
        \label{autoformer_exp}
    }
\end{table}

With the increasing scale of model parameters, large-scale models are becoming increasingly common. 
However, existing NAS algorithms are rarely applied to large vision models. 
To address this gap, we conduct experiments for large-scale transformer-based vision models.
AutoFormer search space is designed for vision transformer architectures and is classified into three levels: tiny, small, and base, facilitating the exploration of more than $1.7 \times 10^{16}$ unique architectures. 
This design facilitates the exploration of architectures with parameter ranges from $4\times10^{6}$ to $75\times10^{6}$.
The results in \Cref{autoformer_exp} demonstrate a significant improvement in accuracy for the generated architectures while achieving a smaller model size, underscoring EDNAG's strong performance and versatility in large-scale vision models. 
Notably, EDNAG achieves significant accuracy improvements across both high-parameter and low-parameter regimes, demonstrating its adaptability to varying computational constraints.

\subsection{Ablation Study}
\label{ablation_studyy}

\begin{figure}[tb]
    \centering
    \includegraphics[width=\linewidth]{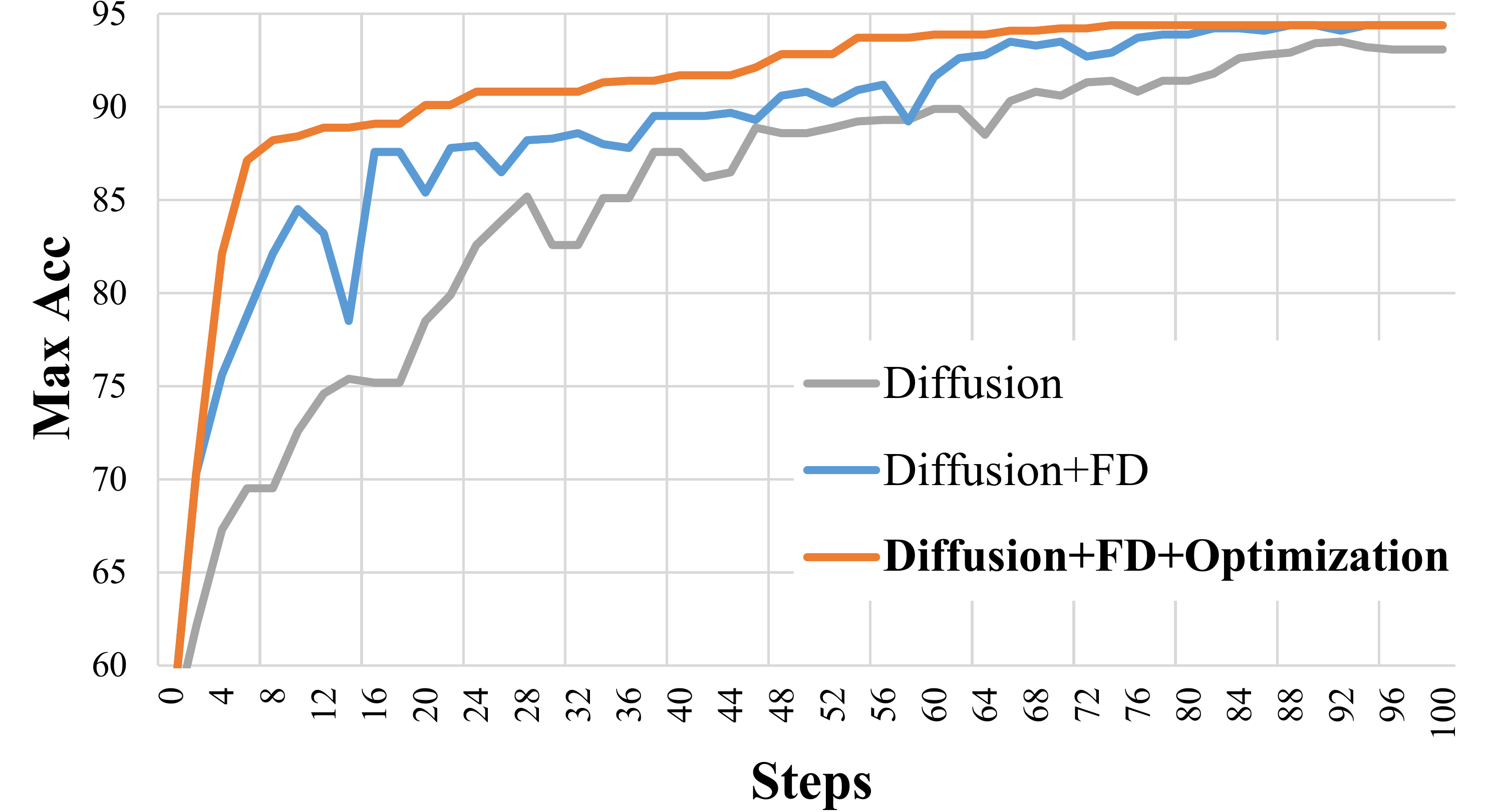}
    \caption{Comparison of architecture generation using Method 1, Method 2, and Method 3. we conducted 3 repeated experiments on CIFAR-10 within NAS-Bench-201.}
    \label{FD_abalation}
\end{figure}
To deeply analyze the contributions of the FD strategy and optimization strategies, we compare architecture generation under three distinct settings: (1) naive diffusion models, (2) diffusion models with the FD strategy, and (3) diffusion models incorporating both FD and optimization strategies, as illustrated in \Cref{FD_abalation}.
Method 1 frequently converges to local optima, yielding suboptimal architectures with an accuracy of $94.04\pm0.16$, while Method 2 notably improves accuracy to $94.34\pm0.02$ by introducing the FD strategy. 
Adding the optimization strategy in Method 3 further stabilizes generation and accelerates convergence, achieving $94.37\pm0.00$ accuracy. 
T-tests reveal statistically significant improvements: $p=0.02334$ (Method 1 vs. Method 3) and $p=0.019$ (Method 2 vs. Method 3). 
These results underscore the effectiveness of the FD strategy and the optimization strategy in EDNAG, demonstrating their pivotal roles in achieving task-optimal architecture generation. 

\subsection{Comparison of Generation Efficiency}


\begin{figure}[tb]
    \centering
    \includegraphics[width=\linewidth]{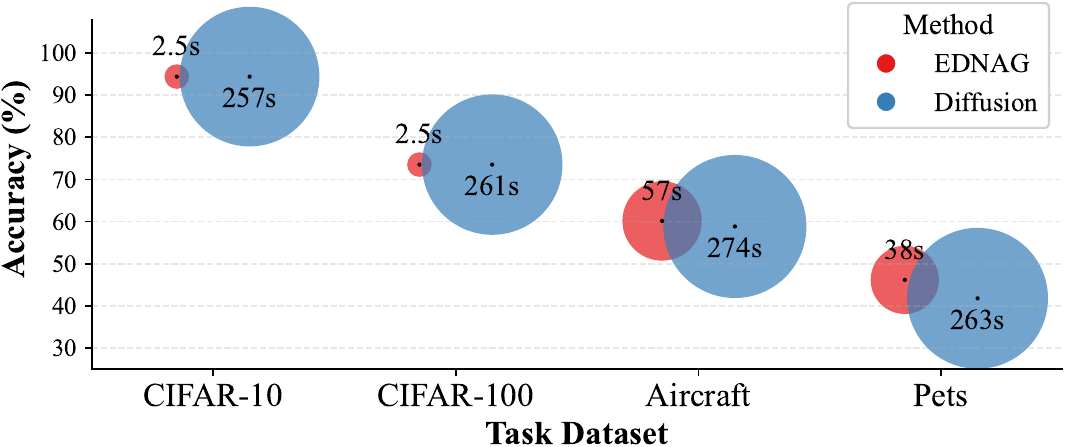}
    \caption{Comparison of GPU time costs. We report inference time for the architecture generation process over 3 runs. Circle radius represents generation time while vertical position indicates accuracies of generated architectures.}
    \label{search_time_EDNAG_DiffusionNAG}
\end{figure}

As a pioneering network-free NAG approach, EDNAG achieves minimal computational resource consumption and rapid conditional generation. 
To demonstrate its efficiency, we compare EDNAG with DiffusionNAG in terms of computational costs during inference in NAS-Bench-201, as shown in \Cref{search_time_EDNAG_DiffusionNAG}. 
Since EDNAG is free of training, it takes \textbf{0} GPU hours for the training process, while DiffusionNAG requires \textbf{3.43} GPU hours on a Tesla V100-SXM2. 
For the generation process, we evaluate the inference time (seconds) of both EDNAG and DiffusionNAG on an NVIDIA RTX 4090. 
EDNAG achieves speedups of up to \textbf{100}$\times$, \textbf{100}$\times$, \textbf{7}$\times$, and \textbf{5}$\times$ compared to DiffusionNAG on CIFAR-10, CIFAR-100, Aircraft and Pets, respectively. 
Therefore, EDNAG enables neural architecture generation without time-consuming training and substantially reduces computational cost during inference. 
In contrast, DiffusionNAG relies on a complex score network at each denoising step, requiring frequent inference and resulting in significantly slower generation.

\section{Conclusion}
This study introduces EDNAG, an efficient and network-free neural architecture generation method.  
We propose a fitness-guided denoising strategy (FD) that maps fitness to a probabilistic space, guiding the DDIM-based denoising process for architecture generation. 
While EDNAG demonstrates strong performance, it remains dependent on a neural predictor for architecture evaluation, making the quality of generated architectures reliant on predictor accuracy.
Future work may address this limitation by replacing the predictor-based evaluation step with evaluation-free strategies (e.g. zero-cost proxies) to guide architecture synthesis more autonomously. 
We believe that EDNAG will contribute to the development of NAS by reducing computational costs and accelerating practical application.

\bibliography{aaai2026}

\begin{thebibliography}{41}
\providecommand{\natexlab}[1]{#1}

\bibitem[{An et~al.(2024)An, Lee, Jo, Lee, and Hwang}]{an2023diffusionnag}
An, S.; Lee, H.; Jo, J.; Lee, S.; and Hwang, S.~J. 2024.
\newblock DiffusionNAG: Predictor-guided Neural Architecture Generation with Diffusion Models.
\newblock In \emph{Proceedings of the 12th International Conference on Learning Representations (ICLR 2024)}.

\bibitem[{Asthana et~al.(2024)Asthana, Conrad, Dawoud, Ortmanns, and Belagiannis}]{asthana2024multi}
Asthana, R.; Conrad, J.; Dawoud, Y.; Ortmanns, M.; and Belagiannis, V. 2024.
\newblock Multi-conditioned Graph Diffusion for Neural Architecture Search.
\newblock \emph{arXiv preprint arXiv:2403.06020}.

\bibitem[{Brock(2018)}]{brock2018large}
Brock, A. 2018.
\newblock Large Scale GAN Training for High Fidelity Natural Image Synthesis.
\newblock \emph{arXiv preprint arXiv:1809.11096}.

\bibitem[{Chen et~al.(2021{\natexlab{a}})Chen, Peng, Fu, and Ling}]{chen2021autoformer}
Chen, M.; Peng, H.; Fu, J.; and Ling, H. 2021{\natexlab{a}}.
\newblock Autoformer: Searching transformers for visual recognition.
\newblock In \emph{Proceedings of the IEEE/CVF international conference on computer vision}, 12270--12280.

\bibitem[{Chen et~al.(2020)Chen, Duan, Chen, Xu, Chen, Liang, Zhang, and Li}]{chen2020catch}
Chen, X.; Duan, Y.; Chen, Z.; Xu, H.; Chen, Z.; Liang, X.; Zhang, T.; and Li, Z. 2020.
\newblock Catch: Context-based meta reinforcement learning for transferrable architecture search.
\newblock In \emph{Computer Vision--ECCV 2020: 16th European Conference, Glasgow, UK, August 23--28, 2020, Proceedings, Part XIX 16}, 185--202. Springer.

\bibitem[{Chen et~al.(2021{\natexlab{b}})Chen, Wang, Cheng, Tang, and Hsieh}]{chen2021drnas}
Chen, X.; Wang, R.; Cheng, M.; Tang, X.; and Hsieh, C.-J. 2021{\natexlab{b}}.
\newblock Dr{NAS}: Dirichlet Neural Architecture Search.
\newblock In \emph{International Conference on Learning Representations}.

\bibitem[{Cowen-Rivers et~al.(2022)Cowen-Rivers, Lyu, Tutunov, Wang, Grosnit, Griffiths, Maravel, Hao, Wang, Peters, and Bou~Ammar}]{cowen2020hebo}
Cowen-Rivers, A.; Lyu, W.; Tutunov, R.; Wang, Z.; Grosnit, A.; Griffiths, R.-R.; Maravel, A.; Hao, J.; Wang, J.; Peters, J.; and Bou~Ammar, H. 2022.
\newblock {HEBO}: Pushing The Limits of Sample-Efficient Hyperparameter Optimisation.
\newblock \emph{Journal of Artificial Intelligence Research}, 74.

\bibitem[{Deng et~al.(2009)Deng, Dong, Socher, Li, Li, and Fei-Fei}]{deng2009imagenet}
Deng, J.; Dong, W.; Socher, R.; Li, L.-J.; Li, K.; and Fei-Fei, L. 2009.
\newblock Imagenet: A large-scale hierarchical image database.
\newblock In \emph{2009 IEEE conference on computer vision and pattern recognition}, 248--255. Ieee.

\bibitem[{Dong et~al.(2024)Dong, Kedziora, Musial, Gabrys et~al.}]{dong2024automated}
Dong, X.; Kedziora, D.~J.; Musial, K.; Gabrys, B.; et~al. 2024.
\newblock Automated deep learning: Neural architecture search is not the end.
\newblock \emph{Foundations and Trends{\textregistered} in Machine Learning}, 17(5): 767--920.

\bibitem[{Dong and Yang(2019)}]{dong2019one}
Dong, X.; and Yang, Y. 2019.
\newblock One-shot neural architecture search via self-evaluated template network.
\newblock In \emph{Proceedings of the IEEE/CVF International Conference on Computer Vision}, 3681--3690.

\bibitem[{Dong and Yang(2020)}]{dong2020bench}
Dong, X.; and Yang, Y. 2020.
\newblock Nas-bench-201: Extending the scope of reproducible neural architecture search.
\newblock \emph{arXiv preprint arXiv:2001.00326}.

\bibitem[{Duan et~al.(2021)Duan, Chen, Xu, Chen, Liang, Zhang, and Li}]{duan2021transnas}
Duan, Y.; Chen, X.; Xu, H.; Chen, Z.; Liang, X.; Zhang, T.; and Li, Z. 2021.
\newblock Transnas-bench-101: Improving transferability and generalizability of cross-task neural architecture search.
\newblock In \emph{Proceedings of the IEEE/CVF Conference on Computer Vision and Pattern Recognition}, 5251--5260.

\bibitem[{Hartl et~al.(2024)Hartl, Zhang, Hazan, and Levin}]{hartl2024heuristically}
Hartl, B.; Zhang, Y.; Hazan, H.; and Levin, M. 2024.
\newblock Heuristically Adaptive Diffusion-Model Evolutionary Strategy.
\newblock \emph{arXiv preprint arXiv:2411.13420}.

\bibitem[{Hemmi, Tanigaki, and Onishi(2024)}]{hemmi2024navigator}
Hemmi, K.; Tanigaki, Y.; and Onishi, M. 2024.
\newblock NAVIGATOR-D3: Neural Architecture Search Using VarIational Graph Auto-encoder Toward Optimal aRchitecture Design for Diverse Datasets.
\newblock In \emph{International Conference on Artificial Neural Networks}, 292--307. Springer.

\bibitem[{Howard et~al.(2019)Howard, Sandler, Chu, Chen, Chen, Tan, Wang, Zhu, Pang, Vasudevan et~al.}]{howard2019searching}
Howard, A.; Sandler, M.; Chu, G.; Chen, L.-C.; Chen, B.; Tan, M.; Wang, W.; Zhu, Y.; Pang, R.; Vasudevan, V.; et~al. 2019.
\newblock Searching for mobilenetv3.
\newblock In \emph{Proceedings of the IEEE/CVF international conference on computer vision}, 1314--1324.

\bibitem[{Huang et~al.(2023)Huang, Wang, Lu, Mohd~Zain, and Yu}]{huang2023scggan}
Huang, Z.; Wang, J.; Lu, X.; Mohd~Zain, A.; and Yu, G. 2023.
\newblock scGGAN: single-cell RNA-seq imputation by graph-based generative adversarial network.
\newblock \emph{Briefings in bioinformatics}, 24(2): bbad040.

\bibitem[{Kang et~al.(2023)Kang, Kang, Kim, Jeon, Chung, and Park}]{kang2023neural}
Kang, J.-S.; Kang, J.; Kim, J.-J.; Jeon, K.-W.; Chung, H.-J.; and Park, B.-H. 2023.
\newblock Neural architecture search survey: A computer vision perspective.
\newblock \emph{Sensors}, 23(3): 1713.

\bibitem[{Lee, Hyung, and Hwang(2021)}]{lee2021rapid}
Lee, H.; Hyung, E.; and Hwang, S.~J. 2021.
\newblock Rapid neural architecture search by learning to generate graphs from datasets.
\newblock \emph{arXiv preprint arXiv:2107.00860}.

\bibitem[{Lee et~al.(2019)Lee, Lee, Kim, Kosiorek, Choi, and Teh}]{lee2019set}
Lee, J.; Lee, Y.; Kim, J.; Kosiorek, A.; Choi, S.; and Teh, Y.~W. 2019.
\newblock Set transformer: A framework for attention-based permutation-invariant neural networks.
\newblock In \emph{International conference on machine learning}, 3744--3753. PMLR.

\bibitem[{Li(2024)}]{li2024theoretically}
Li, G. 2024.
\newblock \emph{Theoretically-grounded efficient deep learning system design}.
\newblock Ph.D. thesis, The University of Texas at Austin.

\bibitem[{Li and Talwalkar(2019)}]{li2020random}
Li, L.; and Talwalkar, A. 2019.
\newblock Random search and reproducibility for neural architecture search.
\newblock In \emph{Uncertainty in Artificial Intelligence}, 367--377. PMLR.

\bibitem[{Lin et~al.(2023)Lin, Zhang, Duan, Ou-Yang, and Zhao}]{lin2023movae}
Lin, Z.; Zhang, Y.; Duan, L.; Ou-Yang, L.; and Zhao, P. 2023.
\newblock MoVAE: a variational AutoEncoder for molecular graph generation.
\newblock In \emph{Proceedings of the 2023 SIAM International Conference on Data Mining (SDM)}, 514--522. SIAM.

\bibitem[{Liu, Simonyan, and Yang(2018)}]{liu2018darts}
Liu, H.; Simonyan, K.; and Yang, Y. 2018.
\newblock Darts: Differentiable architecture search.
\newblock \emph{arXiv preprint arXiv:1806.09055}.

\bibitem[{Liu et~al.(2023)Liu, Gu, Wang, Zhu, Jiang, and You}]{liu2023dream}
Liu, Y.; Gu, J.; Wang, K.; Zhu, Z.; Jiang, W.; and You, Y. 2023.
\newblock Dream: Efficient dataset distillation by representative matching.
\newblock In \emph{Proceedings of the IEEE/CVF International Conference on Computer Vision}, 17314--17324.

\bibitem[{Lopes et~al.(2024)Lopes, Santos, Degardin, and Alexandre}]{LOPES2024127509}
Lopes, V.; Santos, M.; Degardin, B.; and Alexandre, L.~A. 2024.
\newblock Guided evolutionary neural architecture search with efficient performance estimation.
\newblock \emph{Neurocomputing}, 584: 127509.

\bibitem[{Lukasik, Jung, and Keuper(2022)}]{lukasik2022learning}
Lukasik, J.; Jung, S.; and Keuper, M. 2022.
\newblock Learning where to look--generative nas is surprisingly efficient.
\newblock In \emph{European Conference on Computer Vision}, 257--273. Springer.

\bibitem[{Luo et~al.(2024)Luo, Li, Chen, and Zhou}]{luo2024pareto}
Luo, G.; Li, H.; Chen, Z.; and Zhou, Y. 2024.
\newblock Pareto-Informed Multi-objective Neural Architecture Search.
\newblock In \emph{International Conference on Parallel Problem Solving from Nature}, 369--385. Springer.

\bibitem[{Nasir et~al.(2024)Nasir, Earle, Togelius, James, and Cleghorn}]{nasir2024llmatic}
Nasir, M.~U.; Earle, S.; Togelius, J.; James, S.; and Cleghorn, C. 2024.
\newblock LLMatic: neural architecture search via large language models and quality diversity optimization.
\newblock In \emph{proceedings of the Genetic and Evolutionary Computation Conference}, 1110--1118.

\bibitem[{Real et~al.(2019)Real, Aggarwal, Huang, and Le}]{real2019regularized}
Real, E.; Aggarwal, A.; Huang, Y.; and Le, Q.~V. 2019.
\newblock Regularized evolution for image classifier architecture search.
\newblock In \emph{Proceedings of the aaai conference on artificial intelligence (AAAI)}.

\bibitem[{Ru et~al.(2021)Ru, Wan, Dong, and Osborne}]{ru2020interpretable}
Ru, B.; Wan, X.; Dong, X.; and Osborne, M. 2021.
\newblock Interpretable Neural Architecture Search via Bayesian Optimisation with Weisfeiler-Lehman Kernels.
\newblock In \emph{International Conference on Learning Representations}.

\bibitem[{Schulman et~al.(2017)Schulman, Wolski, Dhariwal, Radford, and Klimov}]{schulman2017proximal}
Schulman, J.; Wolski, F.; Dhariwal, P.; Radford, A.; and Klimov, O. 2017.
\newblock Proximal policy optimization algorithms.
\newblock \emph{arXiv preprint arXiv:1707.06347}.

\bibitem[{Shala et~al.(2023)Shala, Elsken, Hutter, and Grabocka}]{shala2023transfer}
Shala, G.; Elsken, T.; Hutter, F.; and Grabocka, J. 2023.
\newblock Transfer {NAS} with Meta-learned Bayesian Surrogates.
\newblock In \emph{The Eleventh International Conference on Learning Representations}.

\bibitem[{Slowik and Kwasnicka(2020)}]{slowik2020evolutionary}
Slowik, A.; and Kwasnicka, H. 2020.
\newblock Evolutionary algorithms and their applications to engineering problems.
\newblock \emph{Neural Computing and Applications}, 32: 12363--12379.

\bibitem[{Song, Meng, and Ermon(2020)}]{song2020denoising}
Song, J.; Meng, C.; and Ermon, S. 2020.
\newblock Denoising diffusion implicit models.
\newblock \emph{arXiv preprint arXiv:2010.02502}.

\bibitem[{Such et~al.(2020)Such, Rawal, Lehman, Stanley, and Clune}]{such2020generative}
Such, F.~P.; Rawal, A.; Lehman, J.; Stanley, K.; and Clune, J. 2020.
\newblock Generative teaching networks: Accelerating neural architecture search by learning to generate synthetic training data.
\newblock In \emph{International Conference on Machine Learning}, 9206--9216. PMLR.

\bibitem[{Vikhar(2016)}]{vikhar2016evolutionary}
Vikhar, P.~A. 2016.
\newblock Evolutionary algorithms: A critical review and its future prospects.
\newblock In \emph{2016 International conference on global trends in signal processing, information computing and communication (ICGTSPICC)}, 261--265. IEEE.

\bibitem[{White, Neiswanger, and Savani(2021)}]{white2019bananas}
White, C.; Neiswanger, W.; and Savani, Y. 2021.
\newblock BANANAS: Bayesian Optimization with Neural Architectures for Neural Architecture Search.
\newblock In \emph{Proceedings of the AAAI Conference on Artificial Intelligence}.

\bibitem[{Xu et~al.(2020)Xu, Xie, Zhang, Chen, Qi, Tian, and Xiong}]{xu2020pc}
Xu, Y.; Xie, L.; Zhang, X.; Chen, X.; Qi, G.-J.; Tian, Q.; and Xiong, H. 2020.
\newblock Pc-darts: Partial channel connections for memory-efficient architecture search.
\newblock In \emph{International Conference on Learning Representations (ICLR)}.

\bibitem[{Zhang et~al.(2019)Zhang, Jiang, Cui, Garnett, and Chen}]{zhang2019d}
Zhang, M.; Jiang, S.; Cui, Z.; Garnett, R.; and Chen, Y. 2019.
\newblock D-vae: A variational autoencoder for directed acyclic graphs.
\newblock \emph{Advances in neural information processing systems}, 32.

\bibitem[{Zhang et~al.(2024)Zhang, Hartl, Hazan, and Levin}]{zhang2024diffusion}
Zhang, Y.; Hartl, B.; Hazan, H.; and Levin, M. 2024.
\newblock Diffusion Models are Evolutionary Algorithms.
\newblock In \emph{International Conference on Learning Representations (ICLR)}.

\bibitem[{Zhou et~al.(2022)Zhou, Sheng, Zheng, Li, Sun, Tian, Chen, and Ji}]{zhou2022training}
Zhou, Q.; Sheng, K.; Zheng, X.; Li, K.; Sun, X.; Tian, Y.; Chen, J.; and Ji, R. 2022.
\newblock Training-free transformer architecture search.
\newblock In \emph{Proceedings of the IEEE/CVF Conference on Computer Vision and Pattern Recognition}, 10894--10903.

\end{thebibliography}

\end{document}